# A Novel Improved Mask RCNN for Multiple Targets Detection in the Indoor Complex Scenes

Zongmin Liu [1,2]*, Jirui Wang [2], Jie Li [2], Pengda Liu [1], Kai Ren [3,4]

1. National Research Base of Intelligent Manufacturing Service, Chongqing Technology and Business University, Chongqing 40006, PR China.
2. School of Mechanical Engineering, Chongqing Technology and Business University, Chongqing 400067, PR China.
3. State Key Laboratory of Fluid Power and Mechatronic Systems, School of Mechanical Engineering, Zhejiang University, Hangzhou, Zhejiang 310027, PR China
4. Key Laboratory of Advanced Manufacturing Technology of Zhejiang Province, School of Mechanical Engineering, Zhejiang University, Hangzhou, Zhejiang 310027, PR China
* Correspondence: liu_zm@ctbu.edu.cn

**Abstract:** With the expansive aging of global population, service robot with living assistance applied in indoor scenes will serve as a crucial role in the field of elderly care and health in the future. Service robots need to detect multiple targets when completing auxiliary tasks. However, indoor scenes are usually complex and there are many types of interference factors, leading to great challenges in the multiple targets detection. To overcome this technical difficulty, a novel improved Mask RCNN method for multiple targets detection in the indoor complex scenes is proposed in this paper. The improved model utilizes Mask RCNN as the network framework. On this basis, Convolutional Block Attention Module (CBAM) with channel mechanism and space mechanism is integrated, and the influence of different background, distance, angle and interference factors are comprehensively considered. Meanwhile, in order to evaluate the detection and identification effects of the established model, a comprehensive evaluation system based on loss function and Mean Average Precision (mAP) is established. For verification, experiments on the detection and identification effects under different distances, backgrounds, angles and interference factors were conducted. The results show that designed model improves the accuracy to a higher level and has a better anti-interference ability than other methods when the detection speed was nearly the same.

**Keywords:** Service robot; Indoor scenes; Multiple targets detection; Mask RCNN; CBAM.

# 1. Instruction

Since the appearance of robot in the 1950s, researchers in various countries have been expanding their research on robots. With the robot technology evolving for half a century, different functional service robots have been developed by scientists and engineers. However, they are unable to enter the majority of users' homes like other household electrical equipment, due to the relevant technology of service robots is not enough mature and perfect. According to the definition of International Federation of Robotics (IFR), service robot regard autonomous or semi-autonomous service function as the core, of which applications mainly include home service, medical assistance, shopping guide, security and etc, except for industrial production [1]. Compared to industrial robots, service robots focus more on the perception and recognition of the environment and objects, as well as the ability to achieve interaction through learning and inference. It is mainly based on the theory of neural networks, optimization algorithms and reinforcement learning, which uses various sensors to achieve autonomous navigation, target detection, task planning, human-computer interaction and other intelligent functions [2-4]. Furthermore, with the increasing of aging population, the research of elderly care and disability assistance robots used in families is an inevitable trend [5]. However, due to the great different indoor scenes and many interferences, it is difficult for robots to achieve the functions of environmental perception, target detection, grasp planning and others, which limits the application and promotion of service robots. Therefore, it is of great theoretical and practical value to take research on accurate detection methods of multiple objects in indoor complex scenes.

As one of the important functions of the service robot, object detection mainly relies on traditional algorithms in early application. For instance, the Histogram of Orientated Gradient (HOG) algorithm was used to achieve pedestrian motion detection [6], and the Deformable Part Model (DPM) algorithm optimized the HOG [7]. With the development of deep learning, AlexNet, which applied ReLU, Dropout and etc to convolutional neural networks (CNN) for the first time, was proposed in 2012 [8], and it shows better performance than traditional algorithms. After that, target detection based on deep learning was mainly divided into two categories: one stage and two stage. One stage relies on target regression detection algorithm, which directly extracts features from the network and predicts target object classification and location. Generally, the detection speed is fast while the accuracy is lower than two stage, such as YOLOv4 [9], EfficientDet [10], YOLOX [11], etc. Comparably, two stage usually adopts the region proposal algorithm. First, the network candidate regions that may contain objects. Then, the second step is to classify samples and regress bounding boxes generated by last step. Normally, the detection accuracy is ideal, but the speed is slower than one stage, such as R-FCN [12], Faster RCNN [13], Cascade RCNN [14], etc. Regions with CNN features (RCNN) method was proposed by Ross Girshick in 2014, and it was the first time to apply CNN to feature extraction achieved better performance than traditional algorithm. Nevertheless, the redundant calculation of overlapping boxes on feature map made the detection speed slow [15]. Based on the framework of RCNN, Faster RCNN removed the traditional selective search method and established a Regional Proposal Network (RPN), which classify and regress the candidate boxes on the feature maps extracted by CNN, thus improving the detection speed and accuracy. Nevertheless, the Regions of Interest (ROI) only considered the last feature map extracted by the backbone network, and it may easily ignore the information of small targets. Furthermore, He et al. [16] improved the basic framework of Faster RCNN and proposed Mask RCNN. This model can not only detect and classify target objects, but also perform instance segmentation, realizing detection tasks accurate to pixel level.

However, considering the diversity of indoor scenes, the application of Mask RCNN in small target detection and interference situation are still limited. In order to expand its scopes of applications, many scholars have proposed improved Mask RCNN models. Typically, literatures [17,18] integrated the attention module, Suppression Network into Mask RCNN to improve performance. They have achieved good results in detecting apples under the conditions of being blocked by leaves, uneven light, background interference and so on, but the detection category was single。Li et al. [19] added a scoring structure to the mask branch of Mask RCNN to detect and classify household common garbage, but such expositions are unsatisfactory for it didn't consider interference like occlusion in the experiment. Hameed et al. [20] proposed a score-based mask edge improvement of Mask RCNN to detect and segment fruits and vegetables in the supermarket. Nonetheless, the experiment results were chiefly single target, and interference factors such as occlusion were not considered. Ma et al. [21] added improved stereo vision algorithm into Mask RCNN to detect and locate fruits in the indoor scenes, providing a reference direction for the research of automatic robot recognition and grasping. Unfortunately, this experiment did not take account of the interference like objects similar with target. Shi et al. [22] removed the mask branch of Mask RCNN and replaced the fully connected layer with Light-Head RCNN to promote the detection speed. This method detected common indoor food and daily necessities with positive effects, but the detection scene was relatively simple and interference factors like occlusion were not taken into consideration. In addition, Zhou et al. [23] adopted Markov model to conduct relational inference for the indoor scenes, which improved the model's generalization ability for object detection. However, the identification in this study were nearly large objects, and no specific detection was carried out for small objects.

With the continuous developing of service robots together with their applications, multi-object detection has become significant in current research particularly in complex scenes of different distances, backgrounds, angles and disturbance factors such as occlusion and similar objects. To overcome these technical problems, based on the principle and behavior of human attention, researchers proposed the theory of attention mechanism, and applied it to enhance neural network's ability of perceiving the important and unimportant parts of data. With the continuous advancement of relevant research, attention mechanisms can be basically divided into two categories: i.e. soft attention and hard attention [24]. Hard attention focuses on the local feature points of the image and the prediction of the highest probability, but it cannot be interpreted by differential theory. In contrast, the core of soft attention is weighted by the average of probabilities, which is differentiable and applicable to the forward propagation and backward feedback of neural network gradients. Therefore, researchers have introduced various soft attention mechanisms into convolutional neural networks one after another. For example, Jaderberg et al. [25] proposed Spatial Transformer Networks (STN), which were used to transform the spatial information of feature maps by scaling, rotating, distorting, and retain the crucial information. Hu et al. [26] designed Squeest-and-Excitation Networks (SENet) to investigate the influence of various feature channels through training for increasing the relevant weights and suppressing the irrelevant weights. Wang et al. [27] believed the dimension reduction of SENet would bring side effects to the prediction of channel attention, so they put forward Efficient Channel Attention (ECA) module. After the global average pooling layer, it used 1x1 convolution layer to replace the original fully connected layer, reducing the amount of parameter calculation and realizing local cross-channel interaction. Nevertheless, the above attention modules only take unilateral feature factors into account, which are difficult to extract more comprehensive output features. Woo et al. [28] presented Convolutional Block Attention Module (CBAM) that connected channel and spatial attention mechanisms, which considered both channel and spatial features comprehensively to achieve good results.

To solve the above problems, firstly, this paper built the indoor scenes dataset with considered the indoor environment of different distance, background, angle, occlusion, and similar interferences, and made the corresponding dataset. Subsequently, data enhancement and transfer learning methods were adopted to retrain on the basis of Mask RCNN pre-trained weight of COCO dataset [29]. Thirdly, CBAM module was combined with Mask RCNN, and it was added into the residual module of backbone to establish the improved Mask RCNN model, so as to upgrade the accuracy and anti-interference ability of target identification in different scenes. Finally, under different backgrounds, distances and various interference factors in indoor scenes we detected multiple types of targets and verify the performance of improved Mask RCNN model. The results show that the improved model carried the accuracy to a higher level and has a better anti-interference ability than other methods when the detection speed is nearly the same.

## 2. Relative work

Mask RCNN network is mostly composed of the following parts: the backbone with inclusion of Resnet [30] and Feature Pyramid Network (FPN) [31], Regional Suggestion Network (RPN), ROI Align, mask predictive branch, and the last full connection layers to achieve the classification and regression. The overall structure is shown in Figure 1.

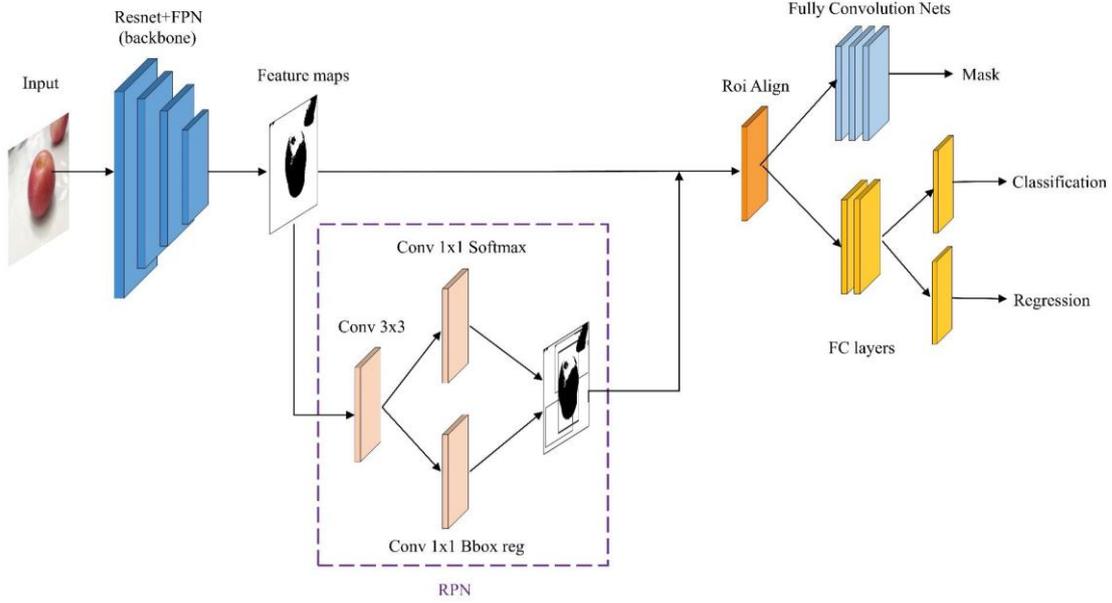

Fig. 1 The structure of Mask RCNN

To balance the detection accuracy and speed in training and testing, Resnet50 is selected for the residual network module, which can effectively solve the problem of gradient disappearance and explosion, and prevent the performance degradation of network. The first layer of Resnet50 is used for processing input pictures while layers 2-5 are composed of basic residual module called bottleneck. The number of bottleneck in layers 2-5 is 3, 4, 6, and 3 in sequence, and the bottleneck structure is shown in Figure 2 (we take the first bottleneck of layer 2 as an example). After Resnet50 down-sampling the input image, the sampled feature maps {C2, C3, C4, C5} is transferred to FPN for up-sampling and multi-scale fusion to obtain {P2, P3, P4, P5} that contains both high-level semantic information and low-level contour information. The general framework is shown in Figure 3, and the calculation formula of the mentioned process defined as:

$$P_5 = Conv_{1 \times 1}(C_5) \tag{1}$$

$$P_i = Conv_{1 \times 1}(C_i) \oplus f_{up} \bullet (P_{i+1}), i \in \{2,3,4\} \tag{2}$$

where $C_i$ is the feature map extracted from layer $i$, and $P_i$ is the feature map through FPN feature fusion of $C_i$. $Conv_{1 \times 1}$ is convolution uses 1x1 kernel, $f_{up}$ is up-sampling, and $\oplus$ is the addition operation of corresponding elements.

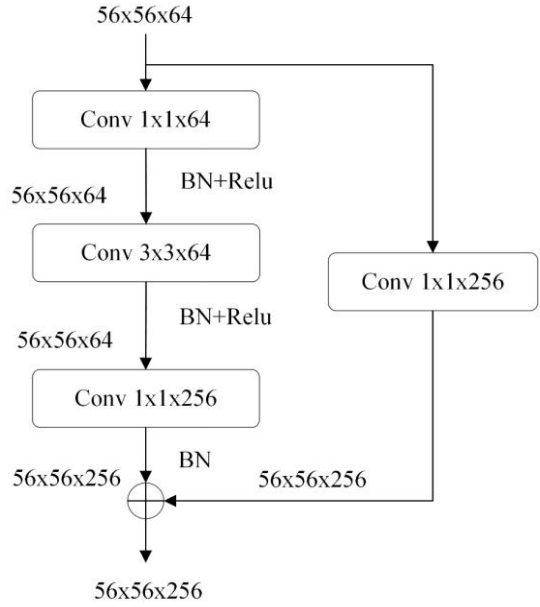

Fig. 2 Bottleneck structure of Resnet50 (layer 2.1)

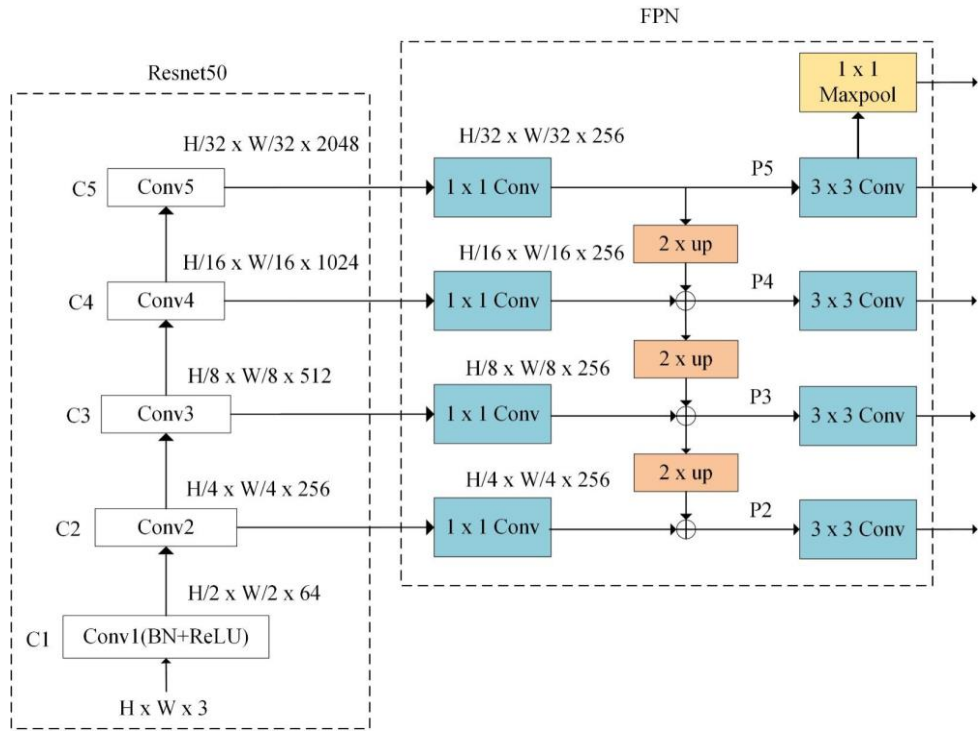

Fig. 3 Backbone network composed of Resnet50 and FPN

The P2-P5 is convolved by 3x3 kernel to reduce the aliasing effects of up-sampling, then transmitted to two branches of RPN. One branch is used to judge the foreground or background, and connected with softmax function for binary classification; Another branch implements bounding box regression, which generates different anchors with different aspect ratios at each pixel of features in all layers, and the aspect ratios are set as [0.5, 1, 2]. Since the size of features in FPN are different, the anchors' area scale are set as $[32^2, 64^2, 128^2, 256^2, 512^2]$, that is, the size of anchor is determined by the located layer. After traversing all pixels, Non-Maximum Suppression (NMS) is adopted to retain anchors with high confidence and get proposals for candidate regions.

In Faster RCNN, the ROI Pooling will pool the corresponding area into a fixed size in the feature map according to the location coordinates of proposals, so as to facilitate subsequent classification and regression operations. As the positions of proposals are obtained by regression, their coordinates are usually not integers. However, the proposals will be quantized twice by ROI Pooling to integer values resulting in an obvious error. ROI Align uses bilinear interpolation instead of quantization to calculate the pixel values of the fixed four positions in each unit. Whereupon, it takes maximum pooling operation effectively to improve the accuracy of output and makes the model more suitable for the detection of small objects.

In addition, Mask RCNN adds a Fully Convolutional Network (FCN), which obtains the exact mask of input image for any size through operations such as convolutional, deconvolution and step-hopping structure to achieve pixel level instance segmentation.

## 3. Method

The development of our method can be divided into three steps: dataset preparing, model building and training and model evaluation. The preparation stage mainly includes dataset production, division, and data augmentation. The training stage includes building the improved Mask RCNN, adopting pre-trained weight for transfer learning, and training the data according to the preset parameters to get the final weight. In the evaluation stage, the performance of model is evaluated by comparing the loss function of training and validating the final weight in the light of measurement. The approximate workflow is shown in Figure 4.

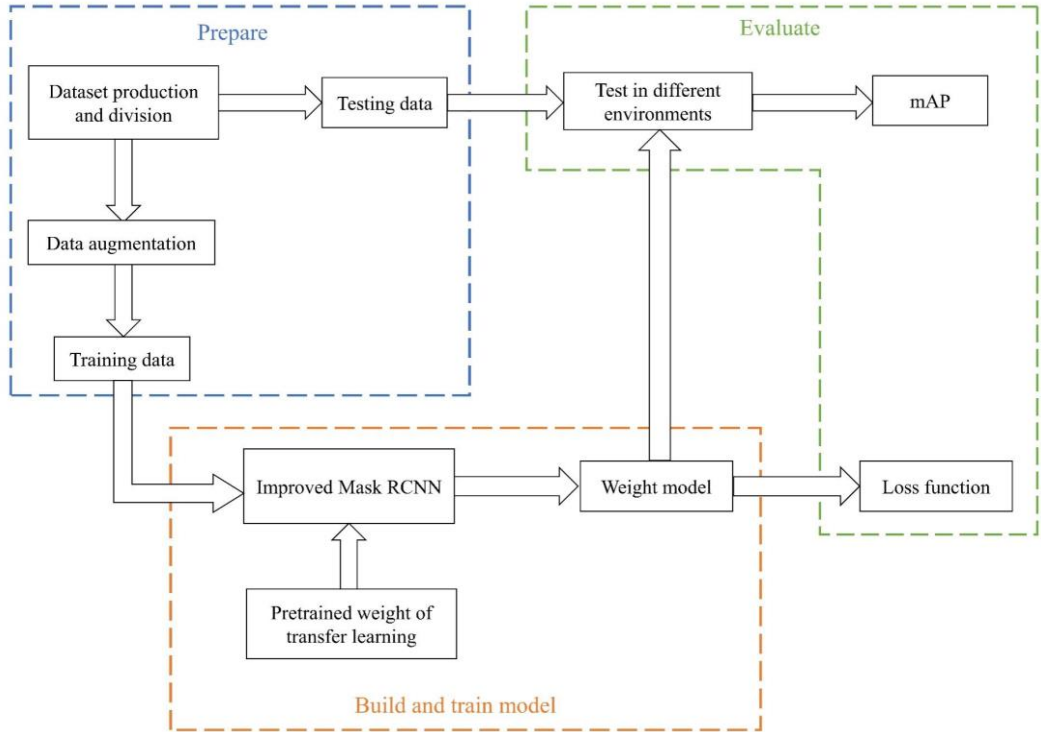

Fig. 4 Workflow

## 3.1 Prepare

During data collection, we kept the indoor light uniform and selected targets include water, cola, BTXL, BHC, JDB, milk, apple, pear and banana. Their sample pictures are shown in Figure 5. Among them, bottled drinks are similar in appearance; JDB and milk are similar in appearance; Apples and pears are similar in appearance; Pears and bananas are similar in color. In addition, our data collection has the following characteristics:

(1) All targets are collected from different backgrounds, angles and distances, as shown in Figure 6:

(2) The picture of dataset not only contain a certain class of targets but also contain two or more classes of targets to ensure the diversity of training data, as shown in Figure 7;

(3) Use data augmentation for some data, including clipping, rotation, flipping and other operations to improve the generalization ability of the training model, as shown in Figure 8.

After making the dataset, we used Labelme tool to annotate and generate json format file for every picture, and convert it into COCO dataset format. The pre-trained weight of

Mask RCNN on official COCO dataset is loaded to actualize transfer learning.

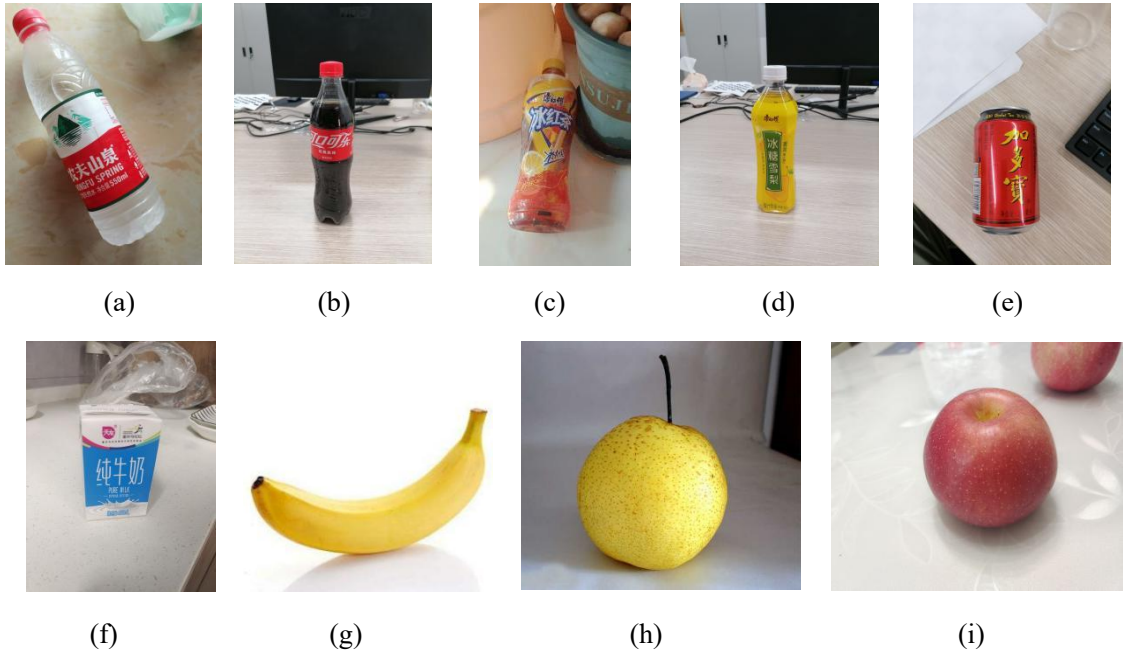

Fig. 5 The categories contained in the dataset: (a) Water. (b) Cola. (c) BHC. (d) BTXL. (e) JDB. (f) Milk. (g) Banana. (h) Pear. (i) Apple

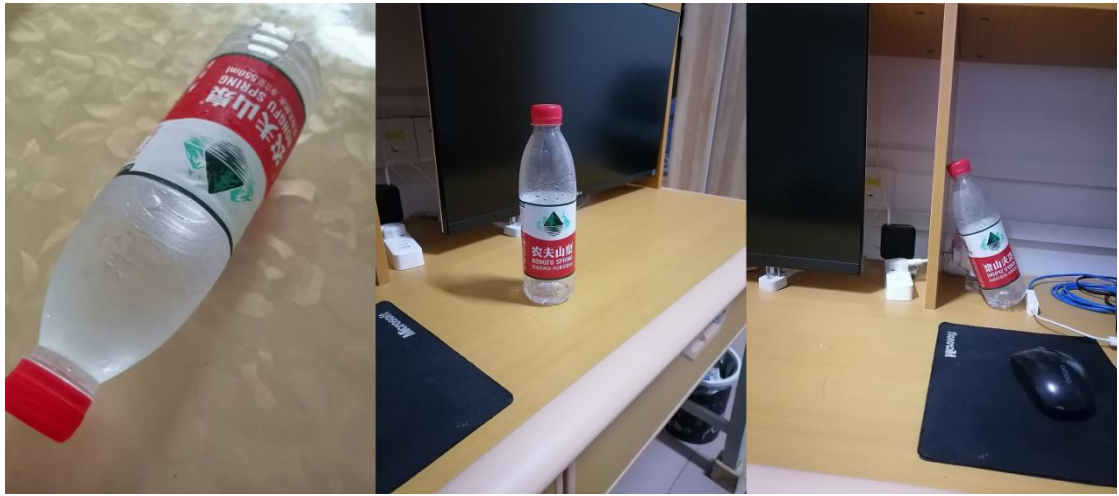

Fig. 6 Data collected from different angles, distances and backgrounds

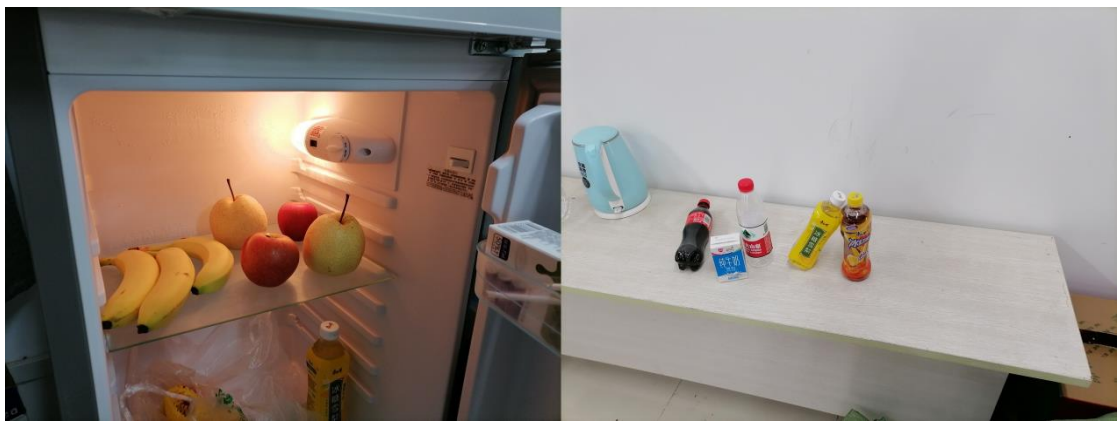

Fig. 7 Containing data for two or more classes of targets

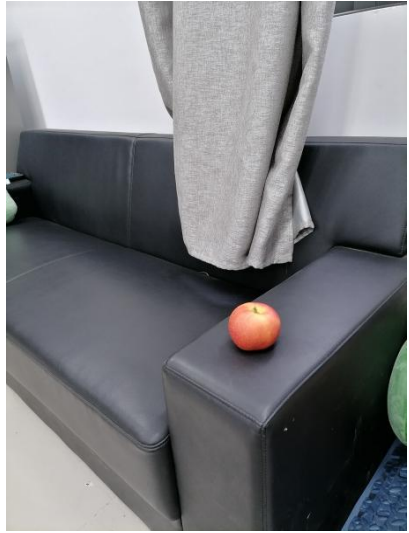

(a)

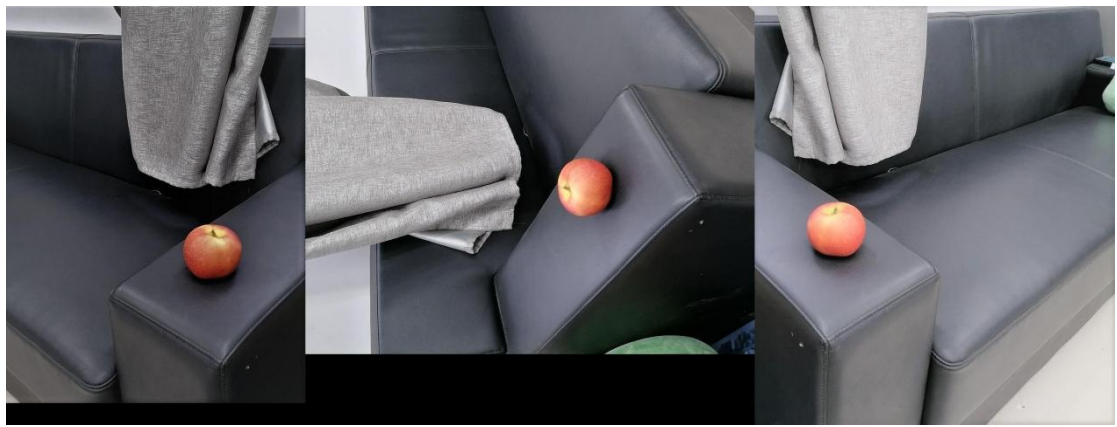

(b)

Figure 8. (a) The original image. (b) The results after data augmentation.

## 3.2 Bulid and train model

Compared with SENet and ECA, CBAM adds Spatial Attention Module (SAM) in addition to Channel Attention Module (CAM), and subsequently connects the two modules in series. It focuses on the specific gravity of channels as well as pixels, so that the model can better suppress the interference information and make the output features are more focused on crucial information. The structure is shown in Figure 9.

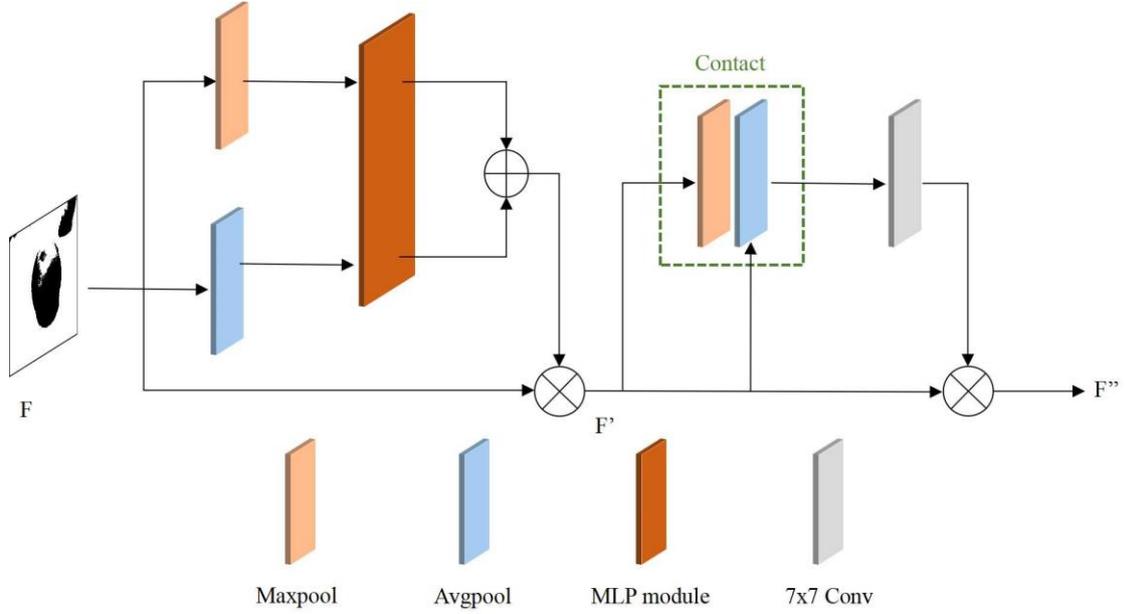

Fig. 9 Structure of CBAM

For input feature $F$ with Channel C, width W and height H, firstly the global maximum pooling based on size, global average pooling and Multi-Layer Perceptron (MLP) with full connection layers were respectively processed in CAM. After that, the corresponding elements of the two matrices were added together and processed by sigmoid function. Finally, multiply the corresponding elements of matrix with the original feature $F$, and the output $F'$ is defined as:

$$F' = F \otimes \sigma(MLP(AP(F)) \oplus MLP(MP(F))) \qquad (3)$$

where $AP$ is global average pooling. $MP$ is global maximum pooling. $MLP$ is Multi-Layer Perceptron operation. $\oplus$ is addition of matrix corresponding elements. $\otimes$ is multiplication of matrix corresponding elements, and $\sigma$ is sigmoid function, defined as:

$$\sigma(x) = \frac{e^x}{e^x + 1} \qquad (4)$$

Input $F'$ into SAM, performing global maximum pooling and global average pooling respectively based on channel, and then contacting the two results to obtain an $H \times W \times 2$ feature. After 7x7 convolution and sigmoid, multiply the corresponding elements with $F'$ to get the output $F''$, which is defined as:

$$F'' = F' \otimes \sigma(Conv_{7\times7}([AP(F'), MP(F')])) \qquad (5)$$

where $Conv_{7\times7}$ represents the convolution of 7x7.

As a plug-and-play module, CBAM can be used in any convolutional layer. In this paper, it is inserted into the end of bottleneck for adequately processing features that build new residue network and obtain the improved Mask RCNN. The advantage of our method is that the attention module acts on every layer to optimize multi-scale features and updates parameters to find the optimal weight matrix in the process of training. For comparison, SENet and ECA insert into the same location, and the modified bottleneck structure is shown in Figure 10.

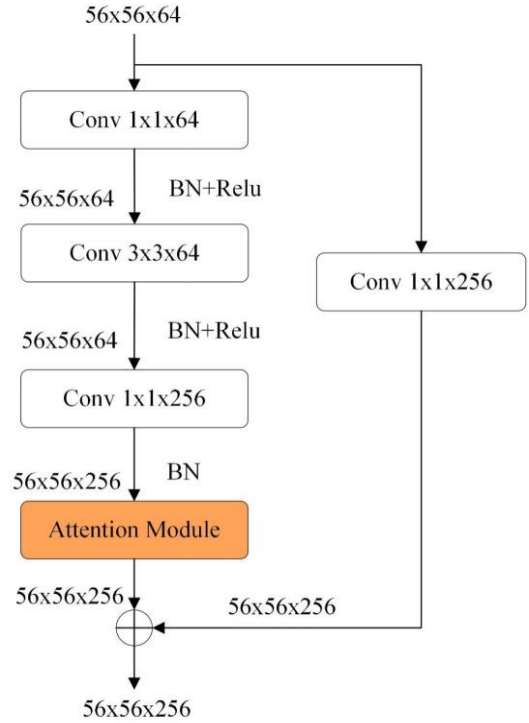

Fig. 10 Bottleneck with attention module

### 3.3 Evaluate

For each ROI, its loss function includes three parts: classified loss $L_{cls}$, regressive loss $L_{reg}$, and mask loss $L_{mask}$. Among them, classified loss is defined as:：

$$L_{cls}(p_i, p_i^*) = -\log[p_i p_i^* + (1-p_i^*)(1-p_i)] \qquad (6)$$

where $i$ is the index of each anchor, $p_i$ is the probability of target that anchor $i$ predicted, and $p_i^*$ is the ground truth label divided into target and background according to the anchor generated by RPN network, defined as:

$$p_i^* = \begin{cases} 0 & otherwise \\ 1 & object\ in\ anchor\ i \end{cases} \quad (7)$$

Regressive loss $L_{reg}$ is defined as:

$$L_{reg}(t_j, t_j^*) = \begin{cases} 0.5(t_j - t_j^*)^2 & if\ |t_j - t_j^*| < 1 \\ |t_j - t_j^*| - 0.5 & otherwise \end{cases} \quad (8)$$

where $t_j = \{t_x, t_y, t_w, t_h\}$ is the offset of anchor $j$ relative to the proposal. $t_j^* = \{t_x^*, t_y^*, t_w^*, t_h^*\}$ is the offset of the anchor $j$ relative to the ground truth, and they are defined as:

$$\begin{cases} t_x = \dfrac{x - x_a}{w_a} & t_y = \dfrac{y - y_a}{h_a} \\ t_w = \log(\dfrac{w}{w_a}) & t_h = \log(\dfrac{h}{h_a}) \end{cases} \quad (9)$$

$$\begin{cases} t_x^* = \dfrac{x^* - x_a}{w_a} & t_y^* = \dfrac{y^* - y_a}{h_a} \\ t_w^* = \log(\dfrac{w^*}{w_a}) & t_h^* = \log(\dfrac{h^*}{h_a}) \end{cases} \quad (10)$$

where $x, y, w, h$ represents the center coordinates, width and height of proposal respectively; $x_a, y_a, w_a, h_a$ are the center coordinates, width and height of current anchor; $x^*, y^*, w^*, h^*$ are the center coordinates, width and height of ground truth.

The mask branch encodes each mask with the size of p*p to generate a binary mask with a resolution ratio of p*p. The mask loss used average binary cross entropy that is defined as follows:

$$L_{mask}(y_{mn}, y_{mn}^*) = -\dfrac{1}{p^2} \sum_{1 \le m, n \le p} [y_{mn}^* \log y_{mn} + (1 - y_{mn}^*)(1 - y_{mn})] \quad (11)$$

where $y_{mn}$ is the prediction result of the central coordinate (m, n) of ROI. $y_{mn}^*$ is the ground truth of (m, n). So the total loss function of the prediction result in our improved Mask RCNN is defined as:

$$L = \dfrac{1}{N_{cls}} \sum_i L_{cls}(p_i, p_i^*) + \dfrac{1}{N_{reg}} \sum_j L_{reg}(t_j, t_j^*) + L_{mask}(y_{mn}, y_{mn}^*) \quad (12)$$

where $N_{cls}$ is the number of ROI, and $N_{reg}$ is the number of pixel points in current feature map.

For object detection model, the evaluation indicators consist of precision, recall, mean average precision (mAP) etc. Generally, according to the relationship between the real category and predicted category, the detection results are divided into four types:

True Positive (TP): both the prediction and ground truth are positive samples;

False Positive (FP): the prediction is a positive sample, but the ground truth is a negative sample;

True Negative (TN): both the prediction and ground truth are negative samples;

False Negative (FN): the prediction is a negative sample, but the ground truth is a positive sample.

Precision (P) refers to the number of real positive samples in the data predicted as positive samples, defined as:

$$P = \frac{TP}{TP+FP} \tag{13}$$

Recall (R) refers to the number of positive samples predicted by the model among the total positive samples, defined as:

$$R = \frac{TP}{TP+FN} \tag{14}$$

Due to precision and recall are two different evaluation indicators, and some models may exist high precision with low recall or vice versa. The P-R curve with precision as ordinate while recall as abscissa considered the two factors comprehensively. The area *AP* (average precision) below the P-R curve is the most commonly used indicator to measure the detection performance of model on targets, and the value is higher, the performance of trained model is better. *mAP* is obtained by averaging the sum of *AP* values of each category that is generally regard as the final evaluation standard of model. The calculation formula is as follows:

$$mAP = \frac{1}{N}\sum_{i}^{N} AP_i \tag{15}$$

where *N* is the number of categories detected.

Intersection over Union (IoU) is the intersection ratio between proposal and ground truth, measuring the accuracy of prediction, defined as:

$$IoU = \frac{proposal \cup groundtruth}{proposal \cap groundtruth} \tag{16}$$

## 4. Experiment and results

### 4.1 Experiment platform

In terms of the experimental platform, Intel RealSense D435i (Figure 11) is selected as the image capture device. It has an RGB-D camera with a resolution of 1280x720. At the same time, the D435i is equipped with a depth sensor, which can obtain the spatial position relationship between camera center and target. This test process was carried out under the operation system Win10. The hardware configuration includes an Intel(R) Core(TM) i9-10900X CPU @ 3.70GHz (20 CPUs), a 12GB GeForce RTX 2080 Ti GPU, and a 64GB RAM. The network was built via the PyTorch [32] deep learning framework for model training and prediction.

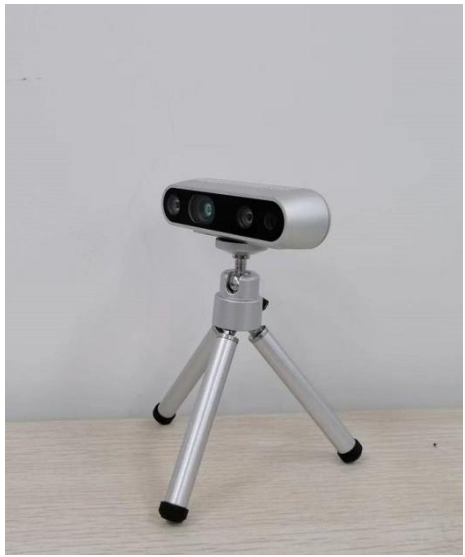

Fig. 11 Intel RealSense d435i

Before model training, the momentum factor was set as 0.9. Learning rate was 0.002. Weight decay was 0.0001 and training times was 26 epochs. Due to the limited GPU memory of equipment, batch size was set as 2 and SGD was selected as the optimization algorithm. The confidence threshold was set as 0.5, and StepLR mechanism was enabled to attenuate the learning rate at the 16th and 22nd epochs while training, and the value after each attenuation was 0.1 times of previous.

## 4.2 Results

In order to verify the performance of the improved Mask RCNN, we selected the original Mask RCNN, Mask RCNN with SENet and Mask RCNN with ECA module as the control groups. To satisfy multiplex requirements, our experiment totally include the different distances and angles between targets and camera, interfering with different backgrounds, colors, appearance and occlusion, and the placement with different posture of targets. In the first group of experiments, all the target objects contained in dataset were arranged on the table, and a relatively close distance was set for detection. The results are shown in Figure 12.

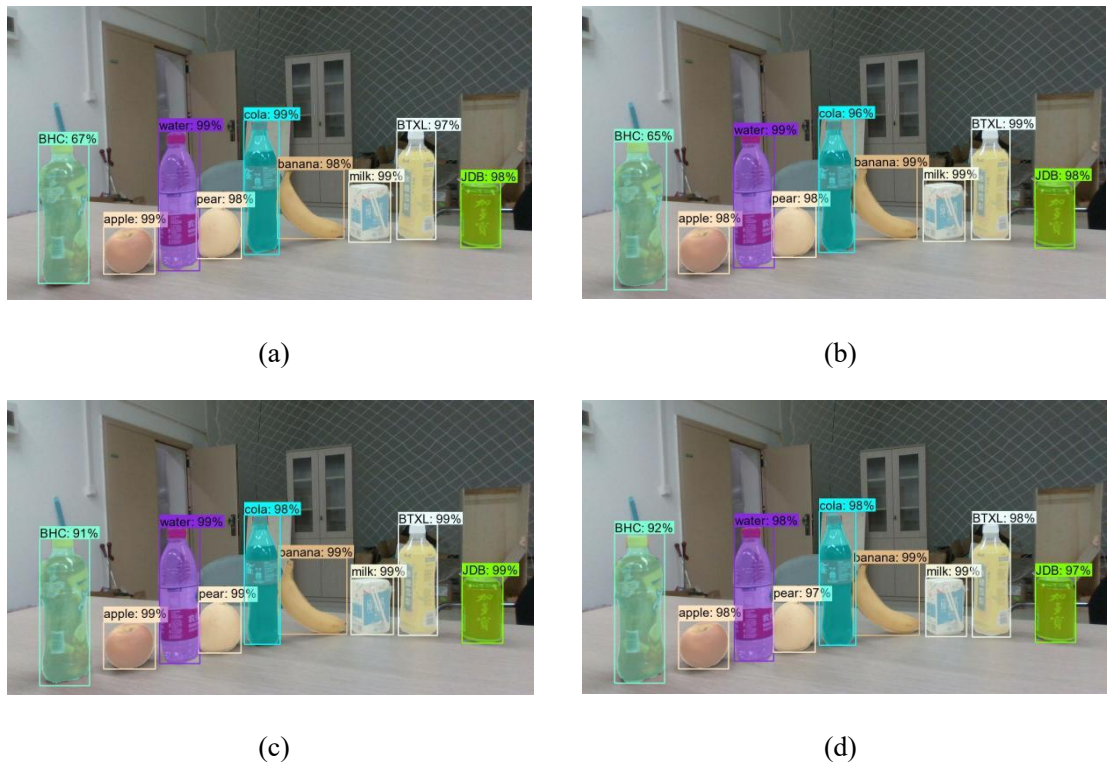

Fig. 12 Detection results at close distance: (a) Mask RCNN; (b) Mask RCNN with ECA; (c) Mask RCNN with SENet; (d) Improved Mask RCNN.

As shown in Figure 12, when the detection results of other targets are almost the same, the "BHC" identified by Mask RCNN and Mask RCNN with ECA have low confidence of 67% and 65% respectively, while the identified by improved Mask RCNN and Mask RCNN with SENet have high confidence (above 90%) that showed better detection effects than former. Second group will change the background and distance to verify the above network. The results are shown in Figure 13.

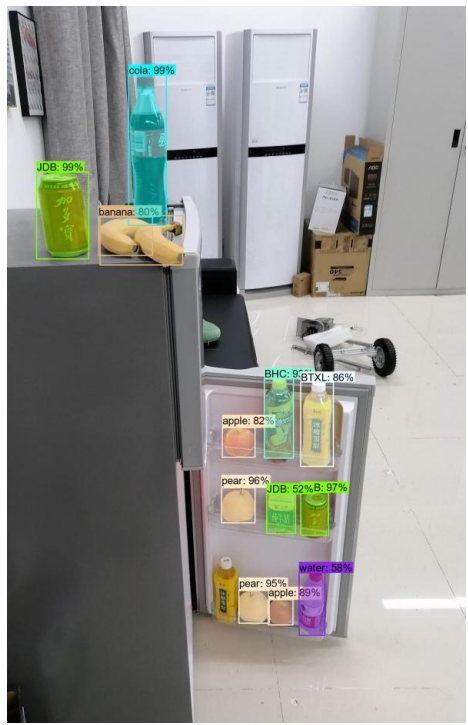
(a)

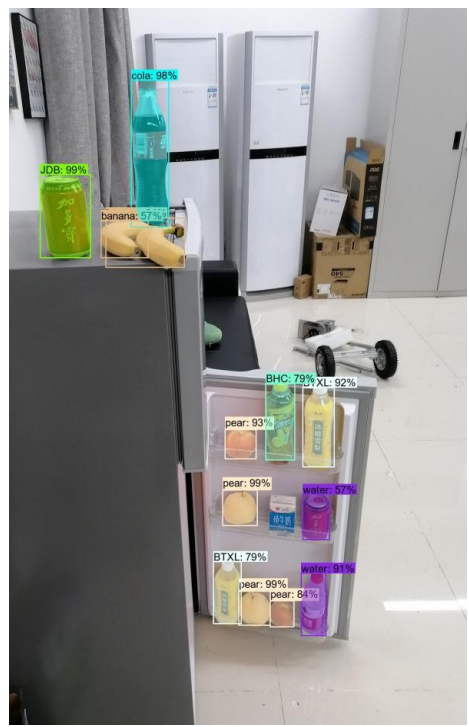
(b)

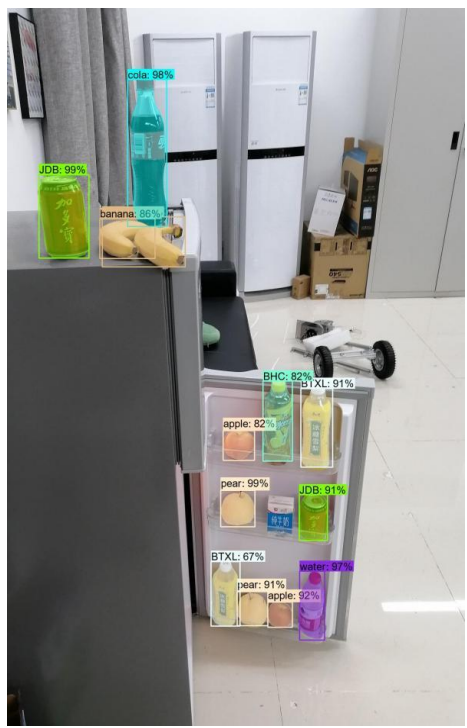
(c)

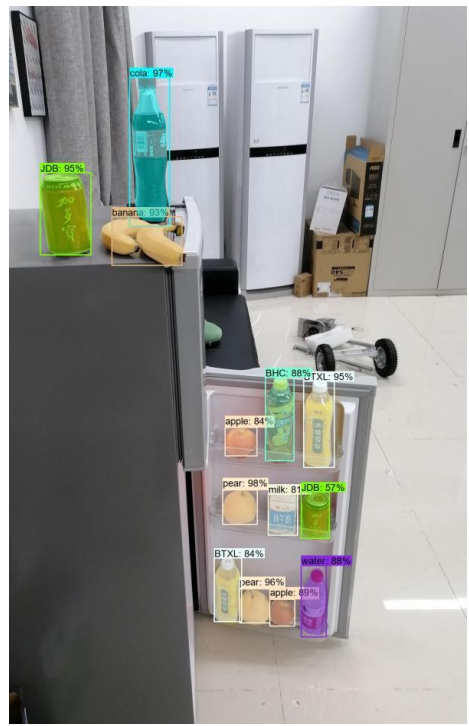
(d)

Fig. 13 Detection results at long distance: (a) Mask RCNN; (b) Mask RCNN with ECA; (c) Mask RCNN with SENet; (d) Improved Mask RCNN.

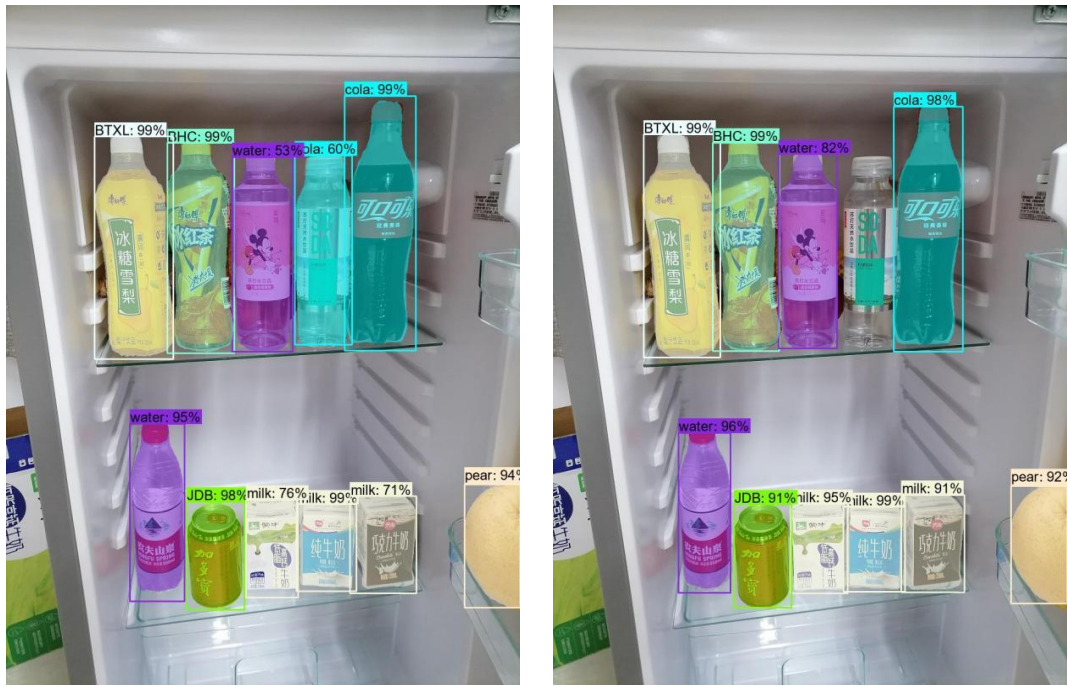

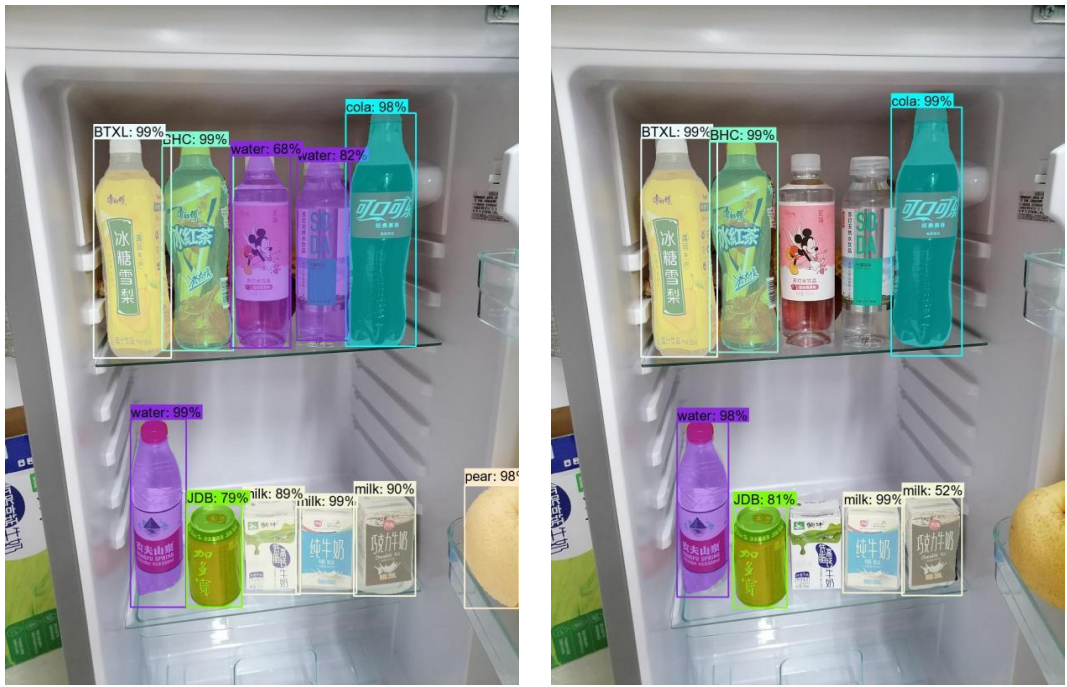

Fig. 14 Detection results under interferences: (a) Mask RCNN; (b) Mask RCNN with ECA; (c) Mask RCNN with SENet; (d) Improved Mask RCNN.

As shown from the Figure 13, improved Mask RCNN can correctly identify all targets in the figure while other models have omissions or errors, indicating the improved model has superior performance than others in small targets detection. In addition, the predicted bounding box of banana in ours has obvious deviation compared with the other three. Considering various shapes of targets, the aspect ratio of anchor could be optimized later and the default value [0.5, 1, 2] can be changed to [1/3, 0.5, 1, 2, 3]. Although the calculated quantity of RPN will be increased, it enhances the model's ability to accurately detect objects with more sizes. At the same time, due to the diversity of family scenes with many interfering factors, the third group will verify the accuracy under the condition of interfering factors, and the results are shown in Figure 14.

Figure 14 shows the ability of distinguish targets and interferences of improved Mask RCNN is significantly advanced compared with other models when there are interferential bottled drinks and boxed milk. Even if the interference is identified as the target milk in (d), the confidence is only 52%. Noticeably, improved Mask RCNN ignores the pear partially occluded in the lower right corner of figure. Given that the indoor scene also exists many conditions that targets are occluded, the fourth group conducted occlusion detection on the targets and the results are shown in Figure 15.

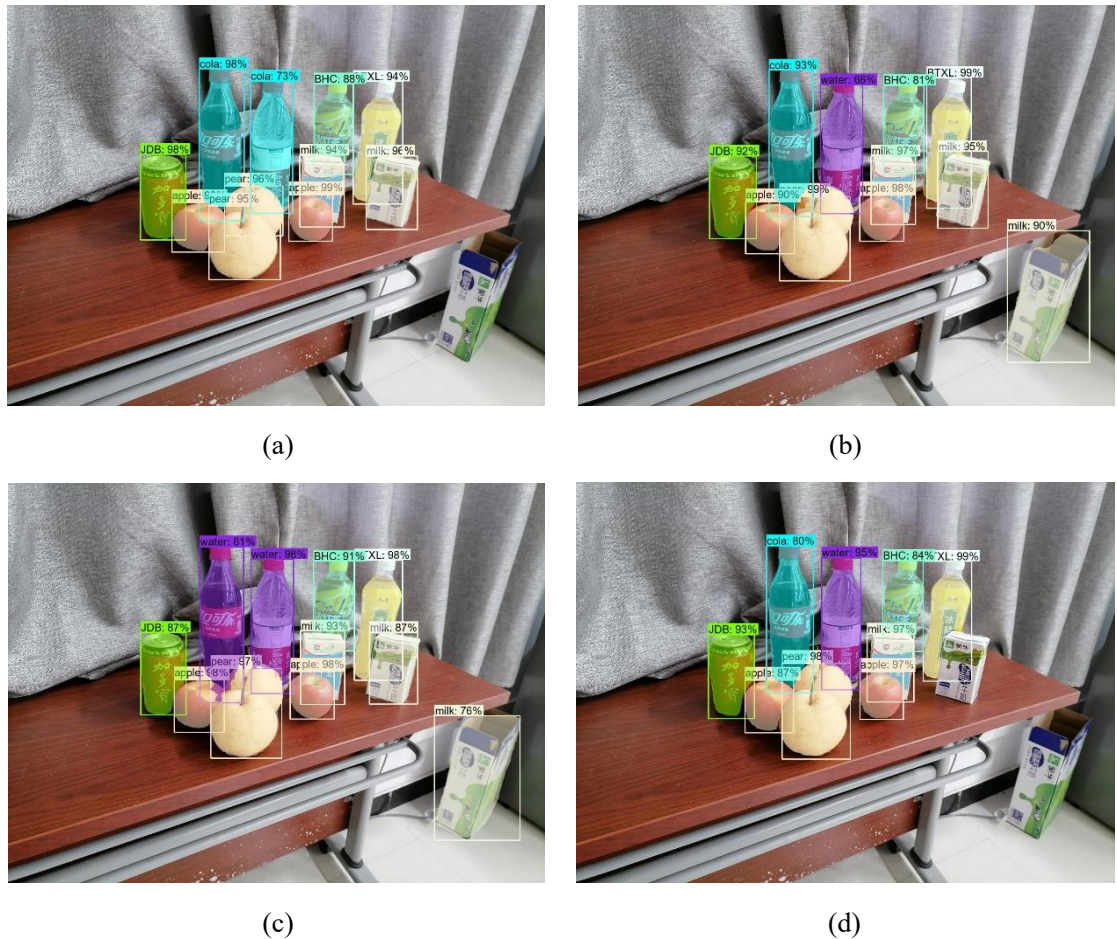

Fig. 15 Detection results under occlusion: (a) Mask RCNN; (b) Mask RCNN with ECA; (c) Mask RCNN with SENet; (d) Improved Mask RCNN.

As shown in Figure 15, the results of four detection models are not satisfying enough. In (a) and (c), there were identification errors of water and cola. In (b) and (c), the box at the bottom right of picture was identified as milk. For the improved Mask RCNN, it's not disturbed by box and interfering milk, which was not achieved by other models, and the confidences of targets are not affected compared with previous groups of experiments. The only flaw in (d) is two pears are identified as one comparing with (a). After analysis, we believe it is because there are a few of occlusion types in training data, and the spatial attention mechanism will increase the weight of related regions between different targets, leading to the poor recognition ability of the overlapped parts. In order to verify the universality of model, next group of experiment we changed the placement of targets, and the results are shown in Figure 16.

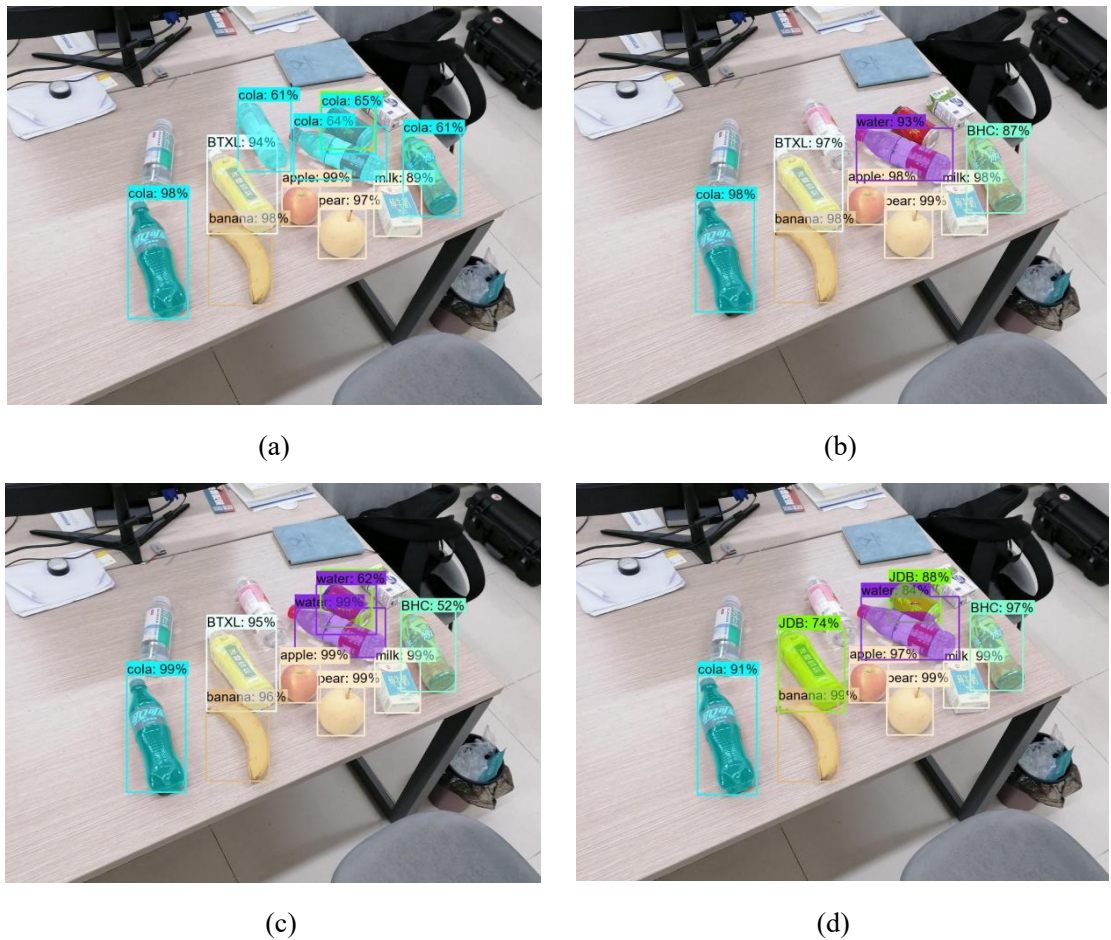

Fig. 16 Detection results with different placement: (a) Mask RCNN; (b) Mask RCNN with ECA; (c) Mask RCNN with SENet; (d) Improved Mask RCNN.

What we acquired from Figure 16 is that improved Mask RCNN still has better anti-interference ability than others for different placement of targets. But the improved model is not perfect for identifying "BTXL" as "JDB", and others also have omissions or errors. After analysis, it is speculated that the training dataset are not rich enough in multifarious poses of targets. In the future, the recognition ability of model will be expanded by increasing the corresponding shapes of data.

The loss function curves derived from the four models after training are shown in Figure 17. It can be seen that the functions are tend to converge in the process of training epochs, and among the three with attention module, the loss of improved Mask RCNN is smaller than other two models, second only to the original Mask RCNN. It is because adding attention modules to bottleneck will increase the training parameters that lead to a little raise of training loss.

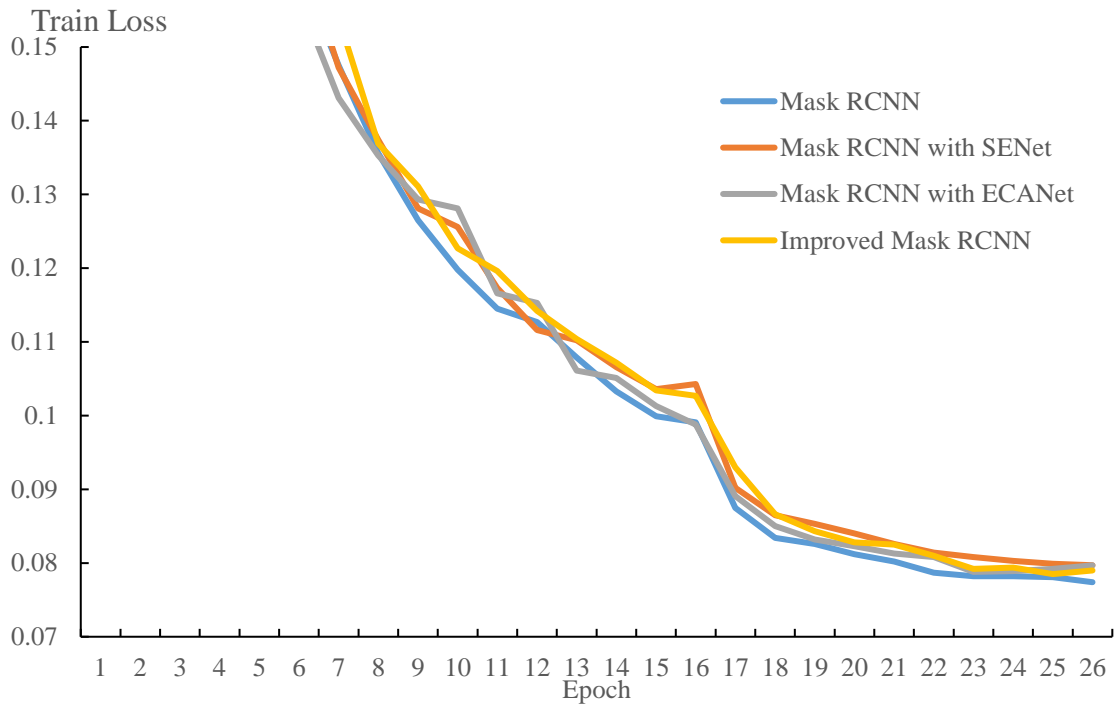

Fig. 17 Loss functions of all models

In terms of evaluation indexes, this paper uses $mAP$ under different IoU for model evaluation, including $mAP^{0.5}$、$mAP^{0.75}$、$mAP^{COCO}$; Where $mAP^{0.5}$ is $mAP$'s value when IoU is 0.5, and $mAP^{0.75}$ is $mAP$'s value when IoU is 0.75. $mAP^{COCO}$ is the evaluation standard of COCO dataset, representing $mAP$'s value when IoU takes the corresponding value every 0.05 increment in the range of [0.50, 0.95], then adapting arithmetic average to obtain the final result. The results of four models are shown in Table 1. It can be seen from the table that improved Mask RCNN has higher value than others obviously in $mAP^{0.5}$ and $mAP^{0.75}$. In $mAP^{COCO}$ it slightly lower than Mask RCNN with SENet, but higher than others.

Table 1 Evaluation Results of Four Models

| Model | $mAP^{0.5}$ | $mAP^{0.75}$ | $mAP^{COCO}$ |
| --- | --- | --- | --- |
| Mask RCNN | 0.9420 | 0.9330 | 0.8219 |
| Mask RCNN with SENet | 0.9582 | 0.9365 | 0.8294 |
| Mask RCNN with ECANet | 0.9514 | 0.9316 | 0.8119 |
| Improved Mask RCNN | 0.9648 | 0.9403 | 0.8231 |

## 5. Conclusion

In this paper, we collected drink and fruit dataset from common indoor scenes, and combined Mask RCNN with CBAM to attained the improved Mask RCNN model. Compared with the original Mask RCNN, Mask RCNN with SENet and Mask RCNN with ECA, our model has the following advantages:

(1) In the feature extraction process, channel and spatial attention mechanisms are integrated, focusing on channel features and spatial features at the same time, further improving the accuracy of target detection in different scenes.

(2) Our model can detect small targets that are difficult to identify by other models in a long distance, and it is more suitable for dynamic changing environment than other methods.

(3) It can distinguish the objects between interference and targets more effectively than other methods, which provided a theoretical reference for the life service robot to recognize and grasp multiple targets in complex indoor scenes.

(4) The loss of our improved model is less than other methods with SENet and ECA, which upgraded the accuracy and ensured the efficiency in the meantime.

However, after several above experiments, there still exist some drawbacks of the model proposed in this paper, such as making a small amount of error under the conditions of occlusion and diverse placement. To solve this problem, we will expand the quantity and type of dataset to make the improved model more adaptable. Moreover, as a two stage network, Mask RCNN spends a lot of time in training and prediction, and requires a large amount of calculation. In the future, we will further optimize the network structure. For example, the algorithm of the one stage network is available on our improved model in this paper to keep accuracy and pick up detection speed. In this way our model will obtain a better practicability in the detection and recognition task of service robot.

**Author Contributions:** Authors' contributions all authors contributed to the study conception and design. Zongmin Liu performed the data analyses and wrote the manuscript. Jirui Wang and Jie Li created the model and performed the experiment. Pengda Liu contributed significantly to analysis and manuscript preparation. Kai Ren helped perform the analysis with constructive discussions. All authors read and approved the final manuscript.

**Funding:** This work was supported by grants of the National Key Research and Development Program of China (No: 2022YFE0107300), the Chongqing Natural Science Foundation (No: cstc2020jcyj-msxmX0067), the Chongqing Technology Innovation and Application Development Special Key Project (No: cstc2021jscx-gksbX0030), the Scientific and Technological Research Program of Chongqing Municipal Education Commission (No: KJQN202000821), the Doctoral Funding of Chongqing University of Chongqing Technology and Business University (No: 1956029), Scientific Research Project of Chongqing Technology and Business University (No: 2152015) and the Graduate Scientific Research and Innovation Foundation of Chongqing Technology and Business University (No: yjscxx2022-112-161).

**Data Availability Statement:** The data that support the findings of this study are available from the corresponding author upon reasonable request.

**Conflicts of Interest:** The authors declare that the research was conducted in the absence of any commercial or financial relationships that could be construed as a potential conflict of interest